# Indoor Positioning System based on Visible Light Communication for Mobile Robot in Nuclear Power Plant

Hongyun Xie [1], Linyi Huang [2], and Wenfei Wu [3]

[1] State Key Laboratory of Nuclear Power Safety Monitoring Technology and Equipment, China Nuclear Power Engineering Co.,Ltd, Shenzhen of Guangdong Prov,518172,China

[2] School of Automation Science and Engineering, South China University of Technology, Guangzhou 510640, China

[3] South-tech Information Technology Co.,LTD, Shenzhen of Guangdong Prov,518172,China

#Hongyun Xie and Linyi Huang should be the co-first author

*Corresponding author: auhly@mail.scut.edu.cn

**ABSTRACT** Visible light positioning (VLP) is widely believed to be a cost-effective answer to the growing demanded for robot indoor positioning. Considering that some extreme environments require robot to be equipped with a precise and radiation-resistance indoor positioning system for doing difficult work, a novel VLP system with high accuracy is proposed to realize the long-playing inspection and intervention under radiation environment. The proposed system with sufficient radiation-tolerance is critical for operational inspection, maintenance and intervention tasks in nuclear facilities. Firstly, we designed intelligent LED lamp with visible light communication (VLC) function to dynamically create the indoor GPS tracking system. By installing the proposed lamps that replace standard lighting in key locations in the nuclear power plant, the proposed system can strengthen the safety of mobile robot and help for efficient inspection in the large-scale field. Secondly, in order to enhance the radiation-tolerance and multi-scenario of the proposed system, we proposed a shielding protection method for the camera vertically installed on the robot, which ensures that the image elements of the camera namely the captured VLP information is not affected by radiation. Besides, with the optimized visible light positioning algorithm based on dispersion calibration method, the proposed VLP system can achieve an average positioning accuracy of 0.82cm and ensure that 90% positioning errors are less than 1.417cm. Therefore, the proposed system not only has sufficient radiation-tolerance but achieve state-of-the-art positioning accuracy in the visible light positioning field.

## I. INTRODUCTION

With the rapid development of nuclear industry, traditional manual maintenance of nuclear equipment cannot avoid radiation harm fundamentally, and manual operation also decreases the safety of nuclear power plant. Therefore, the robot developed for special occasions of nuclear power plant operation, instead of people doing the maintenance, repair and rescue work, can greatly reduce the radiation harm and labor intensity of workers, and has broad application prospects [1]. Different from ordinary robots, nuclear industry mobile robots face even worse environments, such as high radiation, high temperature, high pressure, narrow space, slopes, stairs, and complicated pipeline [2]. Figure 1(a) displays the intelligent robot named as Quince of Japan's Fukushima nuclear power plant [3] and the inspection robot designed by NASA for nuclear power plant is shown in Figure 1(b). These nuclear industry mobile robots play a great role in the process of construction, operation, and maintenance of nuclear power plant. Since the environment becomes even worse and inaccessible after a nuclear accident, robots are more competent than humans to detect signs of life and carry out rescue work in such extreme environment. And in the daily operation of nuclear power plant, the disposal of nuclear waste and the regular inspection and maintenance of facilities are essential and crucial links. However, the work is human-unfriendly, which expose humans to high radiation and may cause a series of diseases. Therefore, as shown in Figure 1(c), various robots [4, 5, 6] designed for inspection and disposal of nuclear waste are used in special areas of nuclear power plant, which not only protect workers from harm of the whole nuclear fuel processing chain, but also fulfill the tasks better. In addition, the previous research on mobile robot system has mainly focused on operational function and position accuracy rather than open management of potential safety hazards and the technical development of robot positioning system under the restrictions of special environment. For example, Quince shown in figure 1(a) abandons the shielding protection for better mobility. Just under such special environmental conditions as nuclear power plant, a high-precision positioning system with sufficient radiation-tolerance is an indispensable basis for inspection robot to complete the tasks safely and smoothly.

FIGURE 1. Typical nuclear industry mobile robot
(a) "Quince" intelligent robot of Japan's Fukushima nuclear power plant
(b) the inspection robot designed by NASA for nuclear power plant
(c) nuclear waste disposing robot and inspection robot

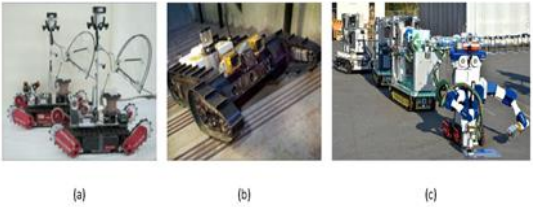

For this purpose, visible light communication (VLC) is a promising method to enhance the robotic system and overcome the environmental limitations of the nuclear power plant. As a wireless transmission technology, VLC uses electrical signals to control the high-speed flashing LEDs to transmit information since the photosensitive device can detect the high-frequency flicker and restore it to the information to be transmitted. VLC paly a great role in increasing the safety of the robot in nuclear power plant, because it can create indoor GPS tracking system through the existing indoor lighting environment and the self-designed smart LED lamps installed in key locations of some special areas to realize the communication between the target environment and the robot positioning system. Traditional positioning technologies such as the positioning technology based on infrared ray or radio frequency (RF) communication [7, 8] often cease to be effective in the face of high-radiation areas or cramped locations in nuclear power plant. Different from conventional indoor positioning technologies mentioned above, visible light positioning (VLP), a kind of positioning technology based on VLC, is not only effective in these scenes, but also superior to other positioning technologies in terms of hardware cost investment and portability. VLP can realize indoor positioning [9, 10, 11] by sending position information of LEDs to the positioning terminal. Because of its high transmission rate and non-electromagnetic interference, VLP has high real-time and multi-scene applicability, which makes the visible light positioning can provide higher accuracy [12].

Therefore, in this paper, we propose a high accuracy visible light positioning system for robot inspection in nuclear power plant, which can overcome the environmental constraints of some special areas such as high radiation, high pressure or high temperature, while maintaining good positioning accuracy within 1cm. In addition, we design the smart LED lamps with VLC function to cooperate with the visible light positioning system. The rest of the paper is organized as follows. Section 2 introduces the design of intelligent LED lamps for VLP, followed by descriptions of the high-precision VLC imaging positioning algorithm based on two LED lamps in Section 3. Then section 4 deeply introduces each part of VLP system. Implementation of the proposed VLP system for inspection robot in nuclear power plant are discussed in section 5, and section 6 is the conclusion.

## II. SELF-DESIGNED INTELLIGENT LED LAMP

### A. HARDWARE DESIGN OF INTELLIGENT LED LAMP

The self-design smart LED lamp is compatible with VLC function with the Bluetooth cloud control model, which makes it easy to change ID code and location-based monitor. The design of smart LED lamp is mainly composed of a power management block and a control block. As shown in Figure 2, the key elements in the power management block are the bridge rectifier, low dropout regulator (LDO) and a LED connected in series. The bridge rectifier and filter capacitor convert 220 V AC input to DC output, so that the lamp can be directly connected with the lamp holder. Then the LDO can generate 5 V stable DC output voltage to supply power to the control block after the input voltage of the LDO is further reduced by the Zener diode connected the positive output terminal of the rectifier. The function of the control block is to generate modulation signals. The control block receives data files and application codes via the low-energy Bluetooth System on Chip (BLE SoC) and stores them in memory. These two blocks are connected by power supply wires and control signal wires. The former works not only as a light source, but also as a DC power supply for the control block. The control block receives data and programs based on Bluetooth, and then generates the modulation signal based on the OOK modulation mode.

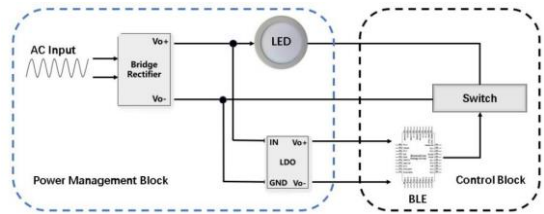

FIGURE 2. Block diagram of intelligent LED lamp

### B. ACTUAL APPLICATION OF INTELLIGENT LED LAMP

Since task implementations and perception feedback relies on reliable communication in the regions, we designed intelligent LED lamp with VLC function to replace standard lighting in key locations in the nuclear power plant, which helps to realize the reliable wireless communication between the proposed system and the target environment. And it is very common for a variety of large fixed equipment exposed to high radiation in nuclear power plant. In order to ensure the normal operation of the equipment, it is necessary to plan to avoid the possible collision between the robot and the equipment. In nuclear power plant, it is generally very difficult to install a physical barrier around large equipment for the purpose of the preventing a robot fall. Since it may cause the potential safety hazards to patrol personnel if a small height difference is installed. Therefore, the safest approach is to warn the robot of the existence of equipment by using a wireless device. However, there are few means to report the information reliably using a wireless system which does not interfere with precision instruments. So robotic system adopted the strategy of warning the inspection robot using VLC. The architecture of an intelligent lighting system is shown in figure 3. Each lamp is assigned a unique LED-ID code that is mapped to the corresponding Bluetooth media access control (MAC) address and location coordinates. The lamps are equipped with an electronic address and have unidirectional communications with the inspection robot. By installing intelligent LED lamps with VLC function around large equipment, in front of stairs and other special areas, the practical management of hazards such as robot dumping can be realized.

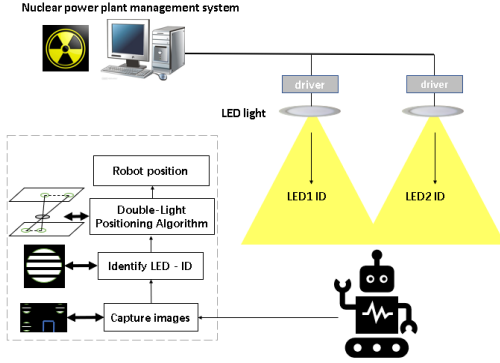

FIGURE 3. Intelligent lighting system architecture using VLC

## III. VISIBLE LIGHT POSITIONING ALGORITHM

This section mainly introduces the positioning algorithm based on double LEDs, because the LED-ID image extraction and recognition have been discussed in our previous work [13]. Generally speaking, visible light positioning first establishes an indoor environment map in order to match the LED-ID with the position coordinate on the map. Formed with light and dark stripes, the image of LEDs can be accurately caught by CMOS camera due to the rolling shutter mechanism. So different lamps can be distinguished in the image since they have distinct stripe features. When two or more LEDs are detected in the vision of the image sensor, we use a relatively compact positioning algorithm, i.e., visible light positioning algorithm. In order to simplify, we only introduce the situation of double LEDs, and the robot selects two appropriate LEDs from the detected LEDs for positioning in reality. As shown in Figure 4, the LEDs are installed on the ceiling, the coordinates of which are $(x_1, y_1, z_1), (x_2, y_2, z_2)$. Since the height of the ceiling is generally the same in one place, the values of z1 and z2 are equal. And (u, v) is the coordinate of LEDs through the Lens in the image coordinate system. In addition, the position of the center point O of the lens can be calculated by the geometric relationship as shown in the Figure 5. According to the relative relationship between the image sensor and the robot base, the coordinate position of the robot can be obtained after a certain coordinate transformation.

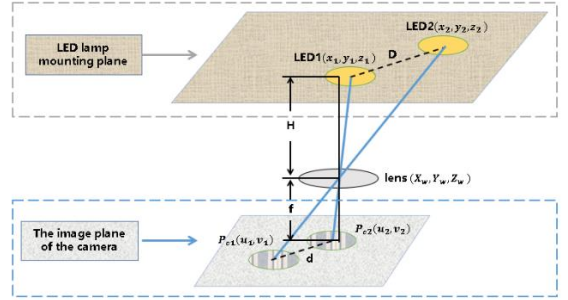

FIGURE 4. The propose visible light positioning system model

Before the coordinate conversion, we need to introduce the relationship between several coordinate systems. Although The image coordinate system and the pixel coordinate system are both located on the imaging plane of image sensor, they have different units of measurement and different origins as seen in Figure 5(a). The origin of the image coordinate system usually defaults to the center of the imaging plane of image sensor. To be more exact, the origin is the intersection point between the optical axis of the camera and the imaging plane of the image sensor. After obtaining the pixel coordinates of LED through camera and image processing, the coordinates of LED in the image coordinate system can be calculated according to the relationship between these two coordinate systems:

$$P_{cn} = (P_{pn} - O_{ps})dl \qquad (1)$$

Where $P_{cn} = (u_n, v_n)$ $(n = 1, 2)$ represents the coordinates of LED in the image coordinate system, $P_{pn} = (i_n, j_n)$ $(n = 1, 2)$ represents the coordinates of LED in the pixel coordinate system, and $O_{ps}$ is the coordinates of the image sensor in the pixel coordinate system, which is thought of the origin of image coordinate system as well as the center of the image. $dl$ represents unit transformation of these two coordinate systems, namely, 1pixel = $dl$ mm. So, the coordinates of the LEDs in pixel coordinate system are transformed into the coordinates in image coordinate system. According to the principle of similar triangles, the coordinate $(x_s, y_s)$ of the terminal installed on the inspection robot in the camera coordinate system can be calculated. If the image coordinate system is parallel to the world coordinate system and they have the same direction, i.e., $(X_c, Y_c, Z_c) = (X_w, Y_w, Z_w)$, the coordinate of the terminal in the world coordinate system is also $(x_s, y_s)$.

$$\frac{x_s - \frac{x_1 + x_2}{2}}{H} = \frac{\frac{u_1 + u_2}{2}}{f} \qquad (2)$$

$$\frac{y_s - \frac{y_1 + y_2}{2}}{H} = \frac{\frac{v_1 + v_2}{2}}{f} \qquad (3)$$

If a robot is moving horizontally on the floor, then only the azimuth angle θ has to be considered and the other two rotation angles are taken to be equal to 0°. In other words, the direction of the world coordinate system and the image coordinate system is not the same more generally, i.e., there is a rotation angle $\theta$ of ZC axis. Hence, the horizontal coordinates of the positioning terminal cannot be obtained directly by the former formula calculation. In this paper, we assumed that the vector from the centroid of the LED A to the centroid of the LED B is parallel to the axis Xc and the direction is the same so that the rotation angle of the image sensor relative to the axis ZC of the world coordinate system can be calculated, which helps to solve the positioning problem in the presence of rotation angle $\theta$. Therefore, as shown in Figure 5(b), the vector from the centroid of B' point in the image corresponding to LED B to the centroid of A' point in the image corresponding to LED A is also parallel to axis XC. Then the rotation angle $\theta$ is equal to the angle between the vector B'A' and the XW axis.

$$\theta = \operatorname{atan2}(v_1 - v_2, u_1 - u_2) \quad (4)$$

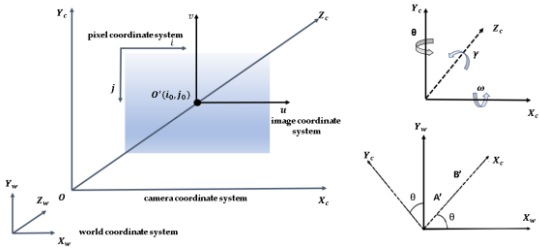

**FIGURE 5.** Transformation between coordinate systems
(a) Relationship between $XcYcZc$ camera coordinate system, $ij$ pixel coordinate system, $uv$ image coordinate system and $XwYwZw$ world coordinate system
(b) Rotation angle of camera coordinate system $(Xc, Yc)$ relative to world coordinate system $(Xw, Yw)$

After accurately identifying LED-ID, LED A and LED B can be determined. Then, we can rotate the coordinates $(x_s, y_s, z_s)$ of the terminal by the follow formula and realize positioning.

$$\begin{bmatrix} X_w \\ Y_w \\ Z_w \end{bmatrix} = \begin{bmatrix} \cos\theta & -\sin\theta & 0 \\ \sin\theta & \cos\theta & 0 \\ 0 & 0 & 1 \end{bmatrix} \begin{bmatrix} X_s \\ Y_s \\ Z_s \end{bmatrix} \quad (5)$$

$(X_w, Y_w, Z_w)$ is the coordinate in the world coordinate system and $(x_s, y_s, z_s)$ is the coordinate in the camera coordinate system. Through the former formula, we can get the value of the coordinate of the inspection robot $(X_w, Y_w)$.

## IV. DESIGN OF THE PROPOSED VLP SYSTEM FOR ROBOT INSPECTION

Founded on the loosely coupled conceptual framework of ROS, the self-designed VLP system can be divided into four parts: VLP information receiver using camera with shielding protection, extraction of LED-ROI by the improved detection algorithm, recognition of LED-ID by machine learning algorithm and calculation of robot position based on visible light positioning algorithm, which integrates separated algorithms to achieve the VLP function and is not affected by electromagnetic interference. Each part is composed of nodes of ROS, which are the minimum unit of executable program that has broken down for the maximum reusability. The communications between nodes such as topic and service are performed by the XML/RPC calls under the TCP/IP protocol so that the nodes can be run in different computers. While the image sensor vertically installed on the robot catches VLP light signal, VLP information can be collected and processed by the proposed algorithms so that the accurate position of robot can be gotten. The detailed process of communication between parts of the VLP system is shown in the Figure 6 and the details of each part are described below:

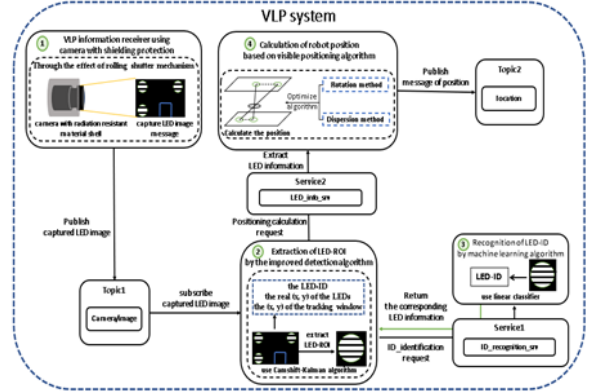

**FIGURE 6.** The structure and communication of VLP system based on visible light positioning algorithm

### A. VLP INFORMATION RECEIVER USING CAMERA WITH SHIELDING PROTECTION

Since the system need to be designed and fabricated to suit the Primary Containment Vessel (PCV) environment: a crowded area with pipes, values, and other fixed structures, the necessary function is considered in this category: a visual inspection system equipped with MindVision camera which is as the interface to receive LED information. And nuclear radiation could affect the normal performance of some parts in the robot, so the design of nuclear protection is also a vital issue, which cannot be ignored. Considering the robot under the high-dose radiation environment, we use cadmium selenium radiation resistant material shell as shielding protection to ensure that the image element of the camera vertically mounted on the robot body is not affected by radiation. When the robot works in the low-dose scenarios, the camera can be just provided simple and thin-walled shields or even abandon the shields for better mobility since the camera has radiation tolerance within a certain dose range. As shown in the corresponding part of figure 6, with the VLP system utilizing the rolling shutter mechanism of CMOS sensors to receive the OOK modulated light signals, the "ON" light signal or the "OFF" light signal can be transformed into bright stripe or dark stripe and then the camera node converts the LED information captured by the CMOS sensors mounted vertically on the robot into ROS image message, which is the basis of realizing the VLP function.

## B. EXTRACTION OF LED-ROI BY THE IMPROVED DETECTION ALGORITHM

It is an efficient method for reducing the computational complexity of image processing to extract the VLP valid information that is LED-ROI. After obtaining the ROS image message, we extract LED-ROI based on pixel intensity detection, which is combined with the dynamic tracking algorithm to reduce the search scope of the next LED-ROI. In order to ensure the accurate detection of LED-ROI at the same time, we choose the automatic threshold method based on Ostu to obtain the good binarization effect without losing information. As described in Ref. [14], the tracking algorithm is improved based on the Camshift-Kalman algorithm. While the Camshift algorithm is applied to track the LED-ROI, the measurement noise of Kalman filter is replaced by the Bhattacharyya coefficient which is the similarity factor between the measurement results of Camshift algorithm and the real target to do better performance on correcting the position of the LED-ROI. And the vector tracking window and the x as well as y coordinates of the tracking window in the image would be stored in a struct. Compared with the original pixel intensity detection, after the initial image frame detects the LED-ROI, the improved detection algorithm can be estimated in the subsequent frames in real time, dynamically and accurately so that the computational time for detecting the precise boundary is neglectable short and the system is more suitable for applying in some complex areas of nuclear power plant.

## C. RECOGNITION OF LED-ID BY MACHINE LEARNING ALGORITHM

This recognition part utilizes machine learning algorithm [15] to recognize the image in LED-ROI by establishing a one-to-one mapping relationship between the characteristics of LED light stripe and LED-ID. And it returns the unique LED-ID to be stored in the mentioned struct together with vector tracking window and its coordinates. In addition, when a LED lamp with VLP function is installed, the LED-ID is associated with the coordinate of the installed LED so that the related LED information is collected enough to calculate the robot position after obtaining the coordinate of the recognized LED.

## D. CALCULATION OF ROBOT POSITION BASED ON VISIBLE LIGHT POSITIONING ALGORITHM

In this part, the position of robot can be computed by the proposed visible light algorithm. If the actual condition is as ideal as we provided, the algorithm will show superior performance of positioning. However, there will inevitably be some unexpected situations in practical application. For example, the camera lens may be incompletely parallel to the image plane in the process of camera manufacturing, which causes the pixel coordinates of the camera are away from the center of the image. Taking such exception into consideration, we proposed two optimized methods, which are rotation method and dispersion method. Through rotating the image sensor around the center of the lens and recording the position of the LED lamp in the image coordinate system every 30 degrees, the actual center coordinates $(u_1, v_1)$ can be obtained. So with using the rotation method, the pixel coordinate system is corrected as follows:

$$u = (i - i_1)du \quad (6)$$
$$v = (j - j_1)dv \quad (7)$$

However, the center coordinates $(u_1, v_1)$ is not enough accurate because we find that a series of recorded position of LED does not completely match a circle. The experimental result shows that there is still error even after correction using rotation method. Based on this situation, we use dispersion method based on the dispersion circle principle to correct the pixel coordinates of the image sensor. In practice, the positioning results are generally symmetrically distributed around the center of the dispersion circle, so that the average value of the recorded results is taken as the position of the center of the dispersion circle for timeliness. We place the image sensor at the origin point of the world coordinate system and make enough measurements to find the smallest circle that can include all the results. The center of the dispersion circle can be represented by the position of the center of the smallest circle containing all the results. Therefore, we can obtain the deviation between the center of the dispersion circle and the actual position, which is shown in the following equation (8) (9):

$$\Delta x = x_c - 0 = \bar{x} = \frac{1}{n}\sum_{i=1}^{n} x_i \quad (8)$$
$$\Delta y = y_c - 0 = \bar{y} = \frac{1}{n}\sum_{i=1}^{n} y_i \quad (9)$$

Based on the deviation, the dispersion method uses the following equation to correct the pixel coordinates of the image sensor:

$$i_1 = i_0 + \frac{\Delta x}{du} \quad (10)$$
$$j_1 = j_0 + \frac{\Delta y}{dv} \quad (11)$$

Combined with the optimized methods, we can get the internal and external parameters of the camera model so that the actual position of camera vertically installed on robot can be calculated as expected.

## V. IMPLEMENTATION OF THE NOVEL VLP SYSTEM

### A. SYSTEM SETUP

All the key system parameters of the proposed VLP based robot system is shown in Table 1. As an experiment platform, mobile robot (TurtleBot3) was used to carry out this VLP system. As shown in Figure 7, we perform the system in the office equipped with several intelligent LED lamps installed on the ceiling for positioning. Clear appearance of LED lamp is also shown in Figure 7. The images of LEDs were shot by MindVision UB-300 industrial camera which is fixed vertically on the mobile robot by prior extrinsic calibration, and transmitted by Raspberry Pi 3 Model B. And the program of image processing and location calculation is run on a remote controller due to the weak processor performance of Raspberry Pi 3 Model B. To ensure the ideal realization of visible light positioning algorithm, the image sensor is supposed to detect at least two LEDs in the positioning process. In addition, to measure the performance of the algorithm, we used Ubuntu 16.04 desktop as the system of remote controller, acer Aspire vn7-593g, Intel (R) Core (TM) i7-7700hq CPU @2.8ghz and Ubuntu 16.04 MATE as software platform, which corresponds to the Kinetic version of ROS.

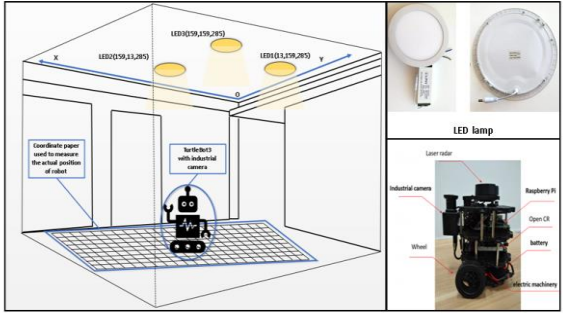

**FIGURE 7.** novel VLP system implement platform

TABLE I
PARAMETERS IN THIS PAPER

| Camera Specifications | |
|---|---|
| Model | MindVision UB-300 |
| Pixel (H × V) | 2048 × 1536 |
| Time of Exposure | 0.02 ms |
| Sensor | 1/2" CMOS |
| Type of Shutter Acquisition Mode | Electronic Rolling Shutter |
| Turtlebot3 Robot Specifications | |
| Module | Raspberry Pi 3 B |
| CPU | Quad Core 1.2 GHz Broadcom BCM2837 |
| RAM | 1 GB |
| operating system | Ubuntu mate 16.04 |
| Remote Controller Specifications | |
| Module | Acer VN7-593G |
| CPU | Quad Core Intel® Core™ i7-7700HQ |
| operating system | Ubuntu 16.04 LTS |
| System Platform Specifications | |
| Size (L ×W× H) | 146 × 146 × 285 cm3 |
| LED Specifications | |
| Coordinates of LED1(cm) | (13, 159, 285) |
| Coordinates of LED2(cm) | (159, 159, 285) |
| Coordinates of LED3(cm) | (159, 13, 285) |
| The half-power angles of LED/deg($\psi_{1/2}$) | 60 |
| Current of each LED | 300mA |
| Rated Power | 18W |
| Optical output | 1500lm @ 6000 K |

## B. EXPERIMENT AND ANALYSIS OF RESULTS

We demonstrate the performance of the proposed VLP system based on ROS, which is designed for robot inspection in nuclear power plant. As described in the section Self-designed VLP system, when VLC program starts, we run the camera node on the mobile robot (TurtleBot3) to capture the images and publish to the "camera/image" topic, which is subscribed by the LED-ROI extraction node. After receiving the images, the LED-ROI extraction node calls for a service to the ID-recognition to get the recognized LEDs information. Finally, with enough VLP information, the position of the mobile robot will be calculated after receiving a location request.

### 1) ERROR ANALYSIS AND COMPARISON BASED ON DIFFERENT OPTIMIZED METHODS

To evaluate the positioning accuracy of the proposed VLP system, two series of experiments were carried out. The optimization performance of the algorithm is verified by comparing the positioning results using the visible light positioning algorithm based on two proposed optimized methods and the actual position of the camera vertically installed on the robot. The latter can be measured by coordinate paper and plumb. And the positioning results can be obtained by running the VLP system mentioned above. As shown in Figure 8, there are altogether 432 (12×36) positioning results in the coordinate paper, of which 36 measuring points are evenly distributed and each position is measured 12 times. The optimized positioning results of robot based on the rotation calibration method are shown in Figure 8(a), and the optimized positioning results of robot based on the dispersion calibration method are shown in Figure 8(b). The black circles represent the actual position of the camera vertically installed on the robot, and the blue crosses represent the positioning results. The distance between them is the positioning error of the VLP system. The experimental results show that the robot position estimated by the proposed algorithm has a high matching degree with the actual position and the proposed algorithm can reach high positioning accuracy. The CDF curves of the two optimizations are shown as Figure 8(c) and (d). And the positioning error histograms based on the two optimization methods are shown in Figure 8(e) and (f). As shown in Figure 8(c), 90% positioning errors of the optimized rotation calibration method are less than 2.140cm, the average positioning error is 1.54cm, and the maximum positioning error is 2.64cm. However, the CDF curves based on the dispersion calibration method in Figure 8(d) show that the 90% positioning error is less than 1.418cm, the average positioning error is reduced to 0.82cm, and the maximum positioning error is less than 1.93cm. And can be seen from the positioning error histograms in Figure 8(e) and (f), it can be obviously found that the peak value of the positioning error corrected by the dispersion method is more concentrated in the area close to 0, while the distribution of the positioning error corrected by the rotation method are more in the outer edge. From the performance of the optimization algorithm, it can be seen that the dispersion method is superior to the rotation method.

In addition, as for the positioning error of the optimized algorithm, some of them are caused by human factors. Firstly, when installing the LED lamp, the position of the LED is not completely what we expected, i.e., there are some deviation in the coordinate of the LED in the world coordinate system. Secondly, when determining the actual position of the camera through the manually drawn network, there will be some errors between the predetermined position and the actual position of the robot. It is inevitable to be confronted with errors in measurement, lighting installation and robot placement. However, our algorithm ensures that the positioning accuracy can be achieved with all experimental errors mentioned above taken into consideration. In conclusion, the dispersion circle optimization algorithm based on double lights has the smallest positioning error. The average positioning error of 0.82 cm and the maximum positioning error of 1.93 cm.

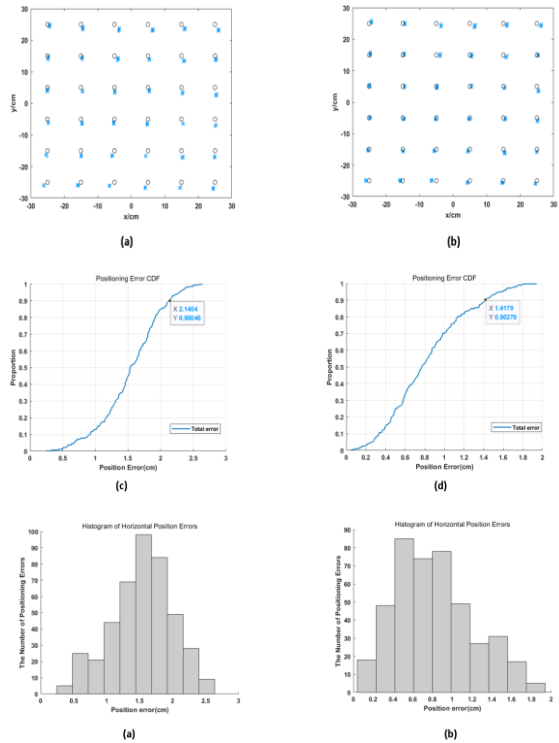

**FIGURE 8.** The positioning results based on two different optimized methods
(a) The positioning results are based on the rotation calibration method.
(b) The positioning results are based on the dispersion calibration method.
(c) The CDF curves of positioning error in visible light rotation optimization positioning system.
(d) The CDF curves of positioning error in visible light dispersion optimization positioning system.
(e) Histogram of the positioning error in visible light rotation optimization positioning system.
(f) Histogram of the positioning error in visible light dispersion optimization positioning system.

### 2) SYSTEM IMPLEMENT AND ANALYSIS BASED ON 3D VISUALIZATION PLATFORM-RVIZ

In the experiment, we used a 3D visualization platform-rviz to display various data, as shown in Figure 9. The middle part of the Figure 9 is the main interface of the robot visualization, showing the map built ahead of time by SLAM, robot and the VLP results. The purple dots in the Figure represent the VLP positioning results of geometric center of the camera vertically installed on the robot, and the displayed information can be controlled by the display setting.

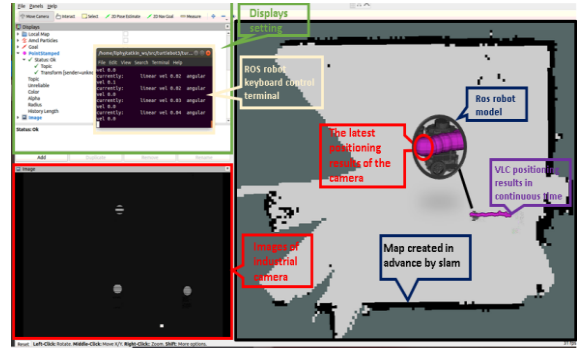

**FIGURE 9.** VLP system for robot inspection in rviz

As shown in figure 10, wherever the robot goes, the purple points representing the positioning results are always perfectly coincident with the robot in the rviz visualization platform. That is to say, the novel visible light positioning system provides a good positioning effect for the robot no matter in the corner, narrow road or the scene with some obstacles. At the same time, it indirectly verifies the feasibility and accuracy of the positioning system in the nuclear power plant with the same environmental complexity. And in Figure 10 (a), the system also shows the sensitivity and robustness of positioning in the transformation of the robot's left and right turning, forward and backward straight travel and other poses, which ensures that the robot has enough flexibility to better realize the dynamic tracking and inspection. In addition, density of the positioning results can reflect that the real-time position of the system and the sequence of purple dots in the Figure 10 (b) almost coincides with the path of the robot in continuous time. In a sense, the system has memory function, which is conducive to timely course correction and route planning.

Based on the above analysis of the positioning results using rviz, we can conclude that the visible light positioning system has a high degree of agreement with the nuclear power plant from the inherent advantages of the system free from electromagnetic interference, its applicability in multiple scenes, real-time positioning and robustness.

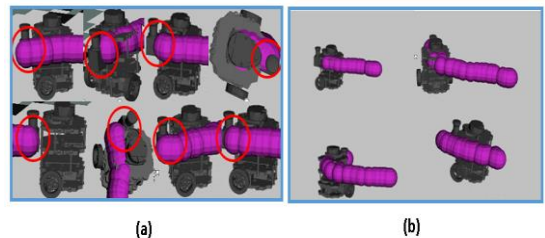

**FIGURE 10.** The positioning effects of VLP system
(a) Positioning effect of robot in various motion states
(b) Positioning effect of robot in continuous time

## V. CONCLUSIONS

This paper presents a centimeter-level novel visible light positioning system with sufficient radiation-tolerance, which is suitable for robot inspection in nuclear power plant. In this paper, the principle of visible light positioning algorithm and the hardware design and actual use of intelligent LED lamp compatible with VLC function in nuclear power plant is deeply and respectively analyzed. The intelligent lamps cooperate with the original indoor lighting system to build the indoor GPS tracking system, which strengthens the safety of mobile robot in the operation of the nuclear power plant. In addition, the VLP system based on ROS has sufficient radiation-tolerance and shows good positioning effect in the experiments of qualitative analysis using 3D visualization platform-rviz and error analysis. Rotating method and dispersion circle method are proposed since the pixel coordinate of the image sensor in the algorithm is not in the center of the image due to the problem of camera manufacturing process. Experiments show that the two methods can optimize the VLP system to a great extent. In optimized double-light positioning algorithm based on dispersion calibration method, 90% positioning errors are less than 1.417cm, and the average positioning error is reduced to 0.82cm, and the maximum positioning error is less than 1.93cm. Thus, the visible light positioning system can be well applied in nuclear power plant due to its sufficient radiation-tolerance, robustness, multi-scenario and real-time positioning.

## REFERENCES

[1] K. Nagatani, S. Kiribayashi, Y. Okada, et al. "Emergency response to the nuclear accident at the Fukushima Daiichi Nuclear Power Plants using mobile rescue robots." Journal of Field Robotics (2013). [2] Qingsong Liu, Guohe Wang, Yachao Dong, Pengfei Liu, Jianquan Sun, Xinyu Wu and Yangsheng Xu, "A novel nuclear station inspection robot." IEEE International Conference on Information Science & Technology IEEE, 2014.

[3] Kinoshita, H., Tayama, R., Kometani, Y., Asano, T. and Kani, Y. (2014), "Development of new technology for Fukushima Daiichi nuclear power station reconstruction", Hitachi Review, Vol. 63 No. 4, pp. 183-190.

[4] Li, I Hsum, W. Y. Wang, and C. K. Tseng. "A Kinect-sensor-based Tracked Robot for Exploring and Climbing Stairs." International Journal of Advanced Robotic Systems 11.1(2014):1.

[5] Ducros, Christian, et al. "RICA: A Tracked Robot for Sampling and Radiological Characterization in the Nuclear Field." Journal of Field Robotics (2016).

[6] Li, Jinke , et al. "A novel inspection robot for nuclear station steam generator secondary side with self-localization." Robotics & Biomimetics 4.1(2017):26.

[7] Yang P., Wu W. Efficient Particle Filter Localization Algorithm in Dense Passive RFID Tag Environment. IEEE Trans. Ind. Electron. 2014, 61, 5641–5651

[8] Yassin, Ali, et al. "Recent Advances in Indoor Localization: A Survey on Theoretical Approaches and Applications." IEEE Communications Surveys & Tutorials PP.99(2016):1-1.

[9] Zhang R, Zhong W D, Qian K, et al. A reversed visible light multitarget localization system via sparse matrix reconstruction[J]. IEEE Internet of Things Journal, 2018, 5(5): 4223-4230.

[10] Xu, Jiaojiao , C. Gong , and Z. Xu . "Experimental Indoor Visible Light Positioning Systems with Centimeter Accuracy Based on A Commercial Smartphone Camera." IEEE Photonics Journal (2018):1-1.

[11] Guan W, Shihuan C, Wen S S, et al. High-Accuracy Robot Indoor Localization Scheme based on Robot Operating System using Visible Light Positioning[J]. IEEE Photonics Journal, 2020.

[12] W. Guan et al., "A novel three-dimensional indoor positioning algorithm design based on visible light communication," Opt. Commun. 392, 282–293 (2017).

[13] C. Xie et al., "The LED-ID detection and recognition method based on visible light positioning using proximity method," IEEE Photonics J. 10(2), 7902116 (2018).

[14] Weipeng Guan. Research on high precision indoor visible light localization algorithm based on image sensor. Master's Thesis, South China University of Technology, Tianhe District and Panyu District of Guangzhou, capital of Guangdong Province, China, 2019

[15] Weipeng Guan, Shangsheng Wen, Lizhao Liu, and Hanlin Zhang "High-precision indoor positioning algorithm based on visible light communication using complementary metal–oxide–semiconductor image sensor," Optical Engineering 58(2), 024101 (8 February 2019).

**Hongyun Xie** is the chief of State Key Laboratory of Nuclear Power Safety Monitoring Technology and Equipment, China Nuclear Power Engineering Co.,LTD. His current research interest is in nuclear industry mobile robot and visible light localization.

**Linyi Huang** is an undergraduate student at South China University of Technology, majoring in automation science and engineering. She will receive the B.S. degrees from South China University of Technology in 2022. Now, her research interests include VLC-based image communication and ROS based robot localization.

**Wenfei Wu** is the CEO of South-tech Information Technology Co.,LTD which is located in Shenzhen of Guangdong Prov in China. His current research interest is in image processing.